\documentclass{article}
\usepackage{spconf,amsmath,graphicx, bbding}
\usepackage[table]{xcolor}
\usepackage{pifont}
\usepackage{booktabs}
\usepackage{svg}

\newcommand{\cmark}{\ding{51}}%
\newcommand{\xmark}{\ding{55}}%

\title{Transfer the linguistic representations from TTS to accent conversion with non-parallel data}
\name{Xi Chen, Jiakun Pei, Liumeng Xue, Mingyang Zhang}
\address{School of Data Science, The Chinese University of Hong Kong, Shenzhen(CUHK-Shenzhen), China}
\begin{document}
%
\maketitle
\begin{abstract}


Accent conversion aims to convert the accent of a source speech to a target accent, meanwhile preserving the speaker's identity. This paper introduces a novel non-autoregressive framework for accent conversion that learns accent-agnostic linguistic representations and employs them to convert the accent in the source speech. Specifically, the proposed system aligns speech representations with linguistic representations obtained from Text-to-Speech (TTS) systems, enabling training of the accent voice conversion model on non-parallel data.
Furthermore, we investigate the effectiveness of a pretraining strategy on native data and different acoustic features within our proposed framework. We conduct a comprehensive evaluation using both subjective and objective metrics to assess the performance of our approach.
The evaluation results highlight the benefits of the pretraining strategy and the incorporation of richer semantic features, resulting in significantly enhanced audio quality and intelligibility~\footnote{Audio samples from this work can be found at https://chenx17.github.io/demos/reference-free-tts-ac/}.

\end{abstract}
\section{Introduction}
\label{sec:intro}


Foreign Accent Conversion (FAC) primarily aims to modify an individual's speech so that it closely resembles a native accent. This process entails adjusting several linguistic elements, such as pronunciation, intonation, and rhythm, while maintaining the identity of the speaker. The accent conversion not only is helpful for pronunciation correction in second-language(L2) learners, but also has potential applications in personalized Text-to-Speech synthesis(TTS)~\cite{Oshima_Takamichi_Toda_Neubig_Sakti_Nakamura_2021}, movie dubbing~\cite{Türk_Arslan_2002}, improving speech recognition performance~\cite{Biadsy_Weiss_Moreno_Kanvesky_Jia_2019}. However, this task remains particularly challenging due to the extreme lack of parallel corpora and the large variations in different speakers. 

The community has witnessed the rapid developement of AC in the past few years. Early works on AC 
focus on \textbf{\textit{reference-based ideas}}~\cite{Zhao_Sonsaat_Levis_Chukharev-Hudilainen_Gutierrez-Osuna_2018, Zhao_Ding_Gutierrez-Osuna_2019, Li_Tang_Yin_Zhao_Li_Wang_Huang_Wang_Ma_2020, ding2022accentron}, these methods typically need target accent speech to provide the reference pronunciation in inference phases, which limits their practical applications.
More recent efforts aim to build a accent conversion system by utilizing the \textbf{\textit{reference-free approaches}}, which can be broadly categorized into two types: \textit{parallel data} based methods~\cite{Zhao_Ding_Gutierrez-Osuna_2021, quamer2022zero, Nguyen_Pham_Waibel} and \textit{non-parallel data} based methods. Approaches belonging to the former type require speech of the same sentence spoken in both the source accent and the target accent simultaneously. Nonetheless, the acquisition of such parallel corpora poses significant challenges, both in terms of cost and feasibility. There are some attempts~\cite{liu2020end,Jin_Serai_Wu_Tjandra_Manohar_He_2022,Zhou_Wu_Zhang_Tian_Li} on learning AC with non-parallel data, the auto-regressive based methods ~\cite{liu2020end,Zhou_Wu_Zhang_Tian_Li} also suffer from the low inference speed and unstable attention, especially when there are multiple speakers.

To tackle the aforementioned issues, we present a novel reference-free non-autoregressive accent conversion framework that learns accent-agnostic linguistic representations and utilize them to convert the accent in the source speech. Specifically, we assume that the linguistic hidden-states are accent-independent, and firstly train a native Fastspeech2~\cite{ren2020fastspeech} model to learn the accent-independent linguistic representations. Furthermore, we use native speech data to learn the alignment between linguistic hidden space and speech hidden space. Finally, we employ accent speech to adapt the alignment module, enabling it to map the accent speech to an accent-agnostic hidden space and drive the decoder to generate native speech mel-spectrogram.

We validate our method by extensive experiments on L2ARCTIC~\cite{zhao2018l2} dataset. The results show that our method achieves competitive performance compared with baseline. The main contribution of our works are three-folds:


$\bullet$ We propose an novel reference-free non-autoregressive accent conversion framework, which leverages the alignment between the speech and text idea to allow the model learned with non-parallel data.
 
$\bullet$ We further explore different input features and pretraining strategy within this framework. The results of experiments demonstrate that these modifications improve audio quality and intelligibility.

$\bullet$ Finally, we conduct comprehensive evaluation, including both objective and subjective metrics, illustrating the effectiveness of our framework.

\section{Method}
\label{sec:format}
Our goal is to build a simple but yet effective accent conversion system. Toward this goal, we develop a reference-free accent conversion approach which doesn't rely on parallel data. In Sec.\ref{sec:problem}, we start with a brief introduction of the accent conversion task formulations mathematically. Subsequently, in Sec.\ref{sec:model}, we detail the proposed model framework. Finally, we describe the training and inference processes in Sec.\ref{sec:traininfer}.

\begin{figure}[tb]
 \centering
 \centerline{\includegraphics[width=0.43\textwidth]{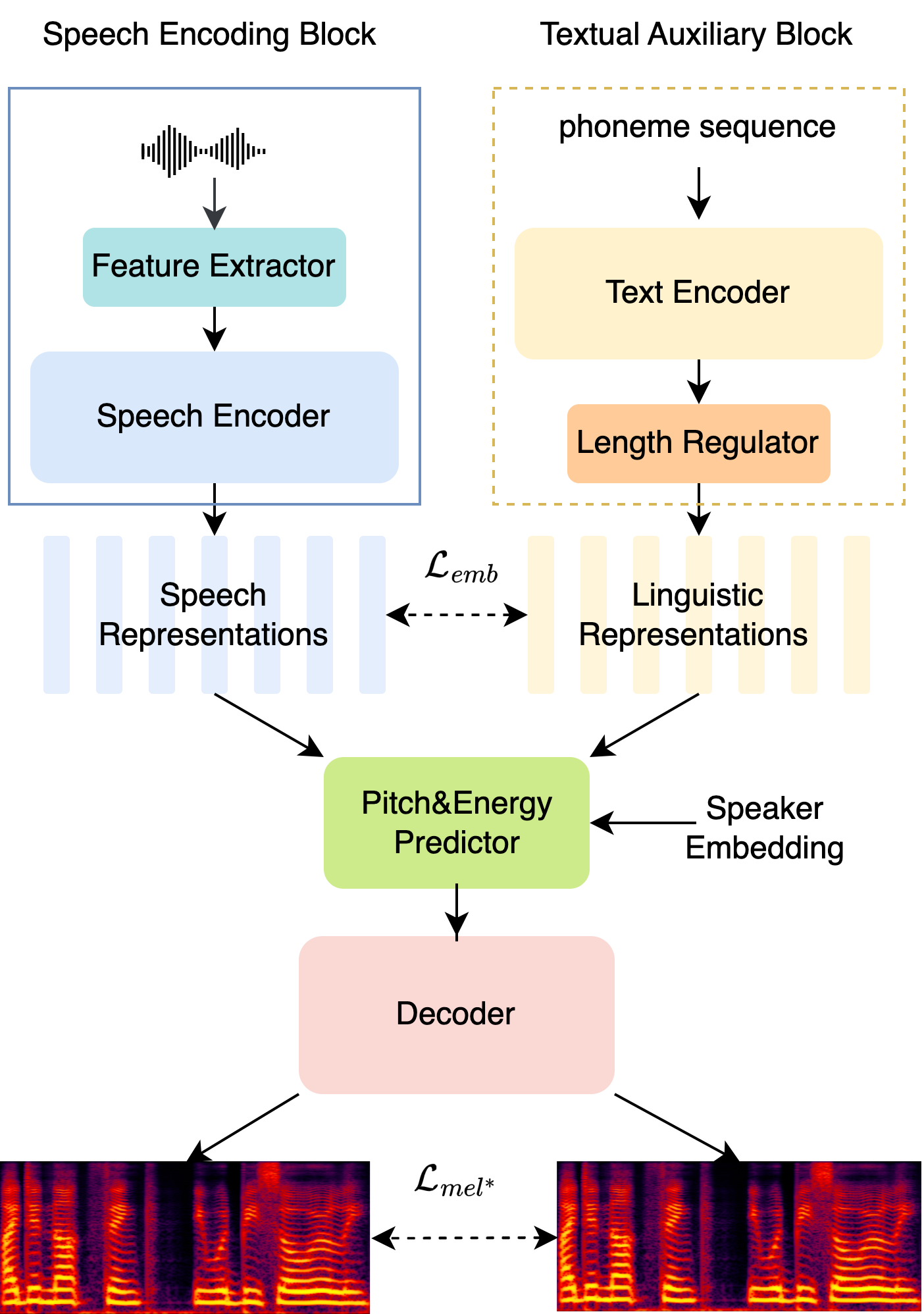}}
\caption{Model Architecture}
\label{fig:arch}
\end{figure}

\subsection{Problem Setting}
\label{sec:problem}
    We now introduce the problem setting of accent conversion. Given a source accent speech, its acoustic features(\textit{e.g.,} mel-spectrogram) are denoted as $\mathbf{A}=\{a_1, a_2, \cdots, a_{T}\}$, $T$ is the timestep of the input features. Accent conversion aims to convert the source speech to the target speech $\mathbf{M}^t=\{m^t_1, m^t_2, \cdots, m^t_{T}\}$ which is spoken in \textbf{target accent}, meanwhile preserve the \textbf{speaker identity}. We can formulate the whole process as:
    \begin{equation}
        \mathbf{M}^t=\mathcal{F}(\mathbf{A})
    \end{equation} Specially, we set the \textit{Hindi accent} is the source accent and the \textit{American accent} is the target accent in this work.

\subsection{Model Architecture}
\label{sec:model}

Our model consists of two encoding blocks together with a decoder block. We use the speech encoding block to extract the speech representations of the source speech. The textual auxiliary block is introduced to help learning the accent-agnostic speech representations, which is only used in training stage.
We instantiate the textual auxiliary block and decoder by adopting the non-autoregressive speech method, FastSpeech2~\cite{ren2020fastspeech}. The framework is illustrated in the Figure~\ref{fig:arch}.

\subsubsection{Speech Encoding Block}
The speech encoding block is compose of the feature extractor and the speech encoder.
Specifically, we have explored different input features for inputs(either \textit{mel-spectrogram} or \textit{whisper feature}). For whisper feature, we adopt the whisper~\cite{radford2023robust} encoder to generate the source speech's acoustic features, denoted $\{a_1, a_2, \cdots, a_{T}\}$.

We follow the previous work and instantiate the speech encoder by stacking the Pre-Net (a linear projection layer) and multiple feed-forward Transformer(FFT) blocks.
We formulate the hidden states extracted from the speech encoder as $\mathbf{H}^s=\{h^s_1, h^s_2, \cdots, h^s_{T}\}$.

\subsubsection{Textual Auxiliary Block}
The textual encoding block is introduced as the auxiliary block to help the disentangled representations learning. In the training stage, we align the the extracted representations of the speech with its linguistic representations generated from the text-encoder followed by the length regulator. 

Formally, we use the phone to represent the source speech's text, and denote the phone as $\mathbf{p}=\{p_1, p_2, \cdots, p_{N}\}$. The latent embedding of the phones generated from text encoder is $\mathbf{E}^l=\{e_1, e_2, \cdots, e_{N}\} $. We then apply the length regulator on the latent embedding to adjust duration, and obtain the length-regulated linguistic representations $\mathbf{H}^l=\{h^l_1, h^l_2, \cdots, h^l_{T}\}$,
$T$ is the length of the input features in time dimension.

By incorporating a textual auxiliary branch, we are able to optimize $\mathbf{H}^s=\{h^s_1, h^s_2, \cdots, h^s_{T}\}$ towards accent-agnostic features $\mathbf{H}^t=\{h^t_1, h^t_2, \cdots, h^t_{T}\}$. Thus, we can decouple the accent from the content in speech, enabling the learning of an accent converter even in the absence of parallel data.

\subsubsection{Decoder}
The decoder is designed to synthesize the converted speech. Given the hidden state representations of the speech, our aim is to generate \textit{speaker-dependent} representations, and synthesize the source speaker's speech with target accent. To accomplish this, we introduce speaker embeddings, which are processed by a pitch and energy predictor, into the existing speech hidden state representations. These speaker-dependent representations then guide the decoder in generating the target accent's mel-spectrograms, denoated as $\mathbf{M}^s$.
The decoder is implemented by stacking multiple feed-forward Transformer(FFT) blocks and PostNet block, as described in~\cite{ren2020fastspeech}. It is worth noting that we employ mel-spectrograms, dented as $\mathbf{M}^t$ decoded from the text input as the reference. 

\subsection{Training and Inference}
\label{sec:traininfer}
\subsubsection{Training}
We split the whole training process into 3 stages to learn a reference-free non-autoregressive accent conversion system.

\noindent \textbf{Stage 1: Learning TTS.}
In this stage, we follow the non-autoregressive method FastSpeech2~\cite{ren2020fastspeech} to learn a TTS model, where the textual auxiliary block , pitch \& energy predictor, and decoder are optimized with native accent data. We denote the loss is $\mathcal{L}_{stage1}$, which includes mel-spectrograms reconstructing loss and duration predictor loss, pitch predictor loss and energy predictor loss.
\begin{equation} \label{eq1}
\mathcal{L}_{stage1} =  \mathcal{L}_{mel}+\mathcal{L}_{duration}+\mathcal{L}_{pitch}+\mathcal{L}_{energy}
\end{equation}

\noindent\textbf{Stage 2: Speech-text Alignment.}
We aims to achieve the alignment between the speech representations and linguistic representations. 
We \textbf{freeze} parameters of all blocks except the \textit{speech encoding block}. We optimize the speech encoder to align speech representations with the linguistic representations via the $\mathcal{L}_{emb}$(\textit{e.g.,} L2 loss). We are able to get a speaker-independent speech encoder in this way.

\begin{equation} \label{eq2}
\begin{split}
\mathcal{L}_{stage2} & = \mathcal{L}_{emb} =  \frac{1}{N} \sum_{i=1}^N\left\|\mathbf{H}^t_i -\mathbf{H}^s_i\right\|_2
\end{split}
\end{equation}

\noindent\textbf{Stage 3: Accent Conversion Fine-tuning.}
Based on the speaker-independent speech encoder, we proceed to fine-tune the speech encoder using \textit{accent data}. The goal is to enable the model to extract accent-independent representations from accented speech inputs. To accomplish this, we employ two loss functions: L2 loss applied to the hidden state space, and L1 loss applied to the mel-spectrograms generated by the decoder. The process can be formally articulated as follows:
\begin{equation} \label{eq3}
\begin{split}
&\mathcal{L}_{stage3}  = \lambda_1 * \mathcal{L}_{emb} + \lambda_2 \mathcal{L}_{mel^*} \\
 &=\lambda_1 * \frac{1}{N} \sum_{i=1}^N\left\|\mathbf{H}^t_i -\mathbf{H}^s_i\right\|_2 + \lambda_2 * \frac{1}{N} \sum_{i=1}^N\left\|\mathbf{M}^t_i -\mathbf{M}^s_i\right\|_1
\end{split}
\end{equation}
$\lambda_1, \lambda_2$ serve as hyper-parameters for the loss functions, $\mathbf{M}^s,\mathbf{M}^t$ represent the mel-spectrograms decoded from speech and text inputs, respectively.

\subsubsection{Inference}
The textual auxiliary block is exclusively employed during the training phases. For inference, the accent conversion system comprises only the speech encoding block, the pitch and energy predictor, and the decoder. Given accented speech, we initially extract its acoustic features using the Whisper model. These acoustic features then serve as input to the speech encoder, which generates accent-independent content features. Subsequently, these speech representations are fused with the speaker embeddings, which are extracted from the source speech. The combined features are then passed through the decoder to produce the converted speech. The resulting output aims to preserve the original voice while adopting a native accent.

\begin{table*}[ht]
\centering
\caption{\textbf{Word Error Rate, MOS(with 95\% confidence interval) and Accentness Results.}}
\label{tab:evalres}
\resizebox{0.8
\textwidth}{!}{%
\begin{tabular}{l|c|c|c|cccc}
\toprule
Name    &     $\mathcal{L}_{mel}$ & WF & Stage2 & WER  & MOS & Accentness\\  
\midrule
Original Hindi Accent & - & - & - &  5.65  &  $4.43\pm0.08$  &  3.49  \\ 
\midrule
TTS(Upper-bound) & - & - & - &  2.04  &  $3.86\pm0.08$  &  1.63 \\ 
\midrule
Baseline & \xmark & \xmark  & \xmark  & 24.59 &  $3.61\pm0.09$  &  1.99  \\ 
Ablation-1 & \cmark & \xmark  & \xmark  & 24.13  & $3.69\pm0.09$ & 1.94 \\  
Ablation-2   & \cmark  & \cmark  & \xmark  & 22.61 & $3.70\pm0.08$ & \textbf{1.86} \\ 
\midrule
\textbf{Our} & \cmark & \cmark & \cmark & \textbf{17.72}  & $\textbf{3.71}\pm0.08$  &  1.90  \\ 
\bottomrule
\end{tabular}}
\end{table*}

\section{Experiments}
\label{sec:pagestyle}
In this section, we conduct a series of experiments to validate the effectiveness of our method. Below we present the experimental configurations in Sec.~\ref{subsec:dataset}, followed by the comprehensive evaluations results in Sec.~\ref{subsec:res}. In addition, we further provide more ablation study and analysis in Sec.~\ref{subsec:ab}.

\subsection{Experimental Configuration}
\label{subsec:dataset}
\textbf{Datasets:} We train the model on the \textit{LibriTTSR}~\cite{koizumi2023libritts} in both stage1 and stage2. It is derived by applying speech restoration to the LibriTTS~\cite{zen2019libritts} corpus, which consists of 585 hours of speech data at 24 kHz sampling rate from 2,456 speakers and the corresponding texts. And then finetune the speech encoder on \textit{L2ARCTIC}~\cite{zhao2018l2}'s Hindi speakers, which are called ASI(M), RRBI(M), SVBI(F), TNI(F). For each speaker, we divided their utterances into three subsets: a training set of 1,032 utterances, a validation set of 50 utterances and a test set of 50 utterances. Specifically, there is no textual overlap among the different sets. A HiFi-GAN~\cite{kong2020hifi} vocoder is used to transform the mel-spectrogram to waveforms.

\noindent\textbf{Configurations:} In stage1 and stage2, we refer to~\cite{vaswani2017attention} used the Adam optimizer with $\beta_1=0.9, \beta_2=0.98$ and $\epsilon=10^{-9}$. We use warm-up for lr scheduler and set $warmup\_steps=4000$.
Model achieves the convergence after 100000 steps and 200000 steps in stage1 and stage2, respectively. Moreover, we use a small learning rate $lrate=10^{-5}$ in stage3, to adapt the speech encoding block. We refer the reader to the supplementary material for the detailed configurations.

\noindent\textbf{Subjective Evaluation Metrics.}
For subjective evaluation, we have conducted three tests, including \textit{acoustic quality test}, \textit{voice identity test}, and \textit{accentness test}.
In the Acoustic Quality Test, a subset of 20 utterances was randomly selected from the test corpus. 20 participants was then enlisted to evaluate the acoustic quality of each utterance. Evaluations were conducted using a 5-point Mean Opinion Score (\textbf{MOS}) as the standard rating metric. For the Voice Identity Test, we performed speaker similarity preference assessments by comparing the converted outputs from various systems against the original utterances with an Hindi accent. In this test, 10 utterances were randomly drawn from the test corpus, and 20 participants was asked to indicate their preferred utterance based on speaker similarity. In the Accentness Test, the experimental setup was analogous to that of the Acoustic Quality Test. Participants were instructed to rate the degree to which the converted utterance retained the source accent. Ratings were made on a 5-point scale, where a score of 1 indicated "No Accent" and a score of 5 indicated a "Strong Accent."

\noindent\textbf{Objective Evaluation Metrics.}
    As for objective metrics, \textit{Word Error Rate (WER)} was utilized to quantify the difference between a reference transcription and a hypothesis transcription generated by a competitive ASR system~\cite{radford2023robust}.


\begin{figure}[tb]
  \centering
  \centerline{\includegraphics[width=0.45\textwidth]{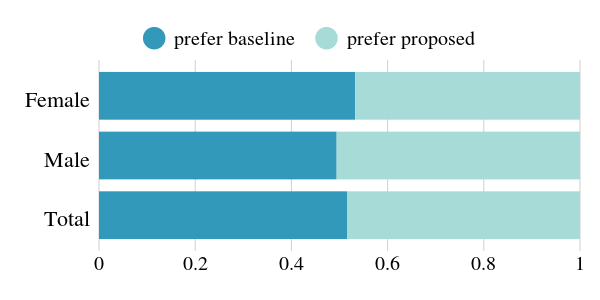}}
\caption{Speaker similarity preference test results.}
  \label{fig:speaker_prefer}
\end{figure}

\subsection{Evaluation Results: Baseline \textit{v.s.} Our Method}
\label{subsec:res}
We employ a multi-dimension evaluation that incorporates both objective metrics and subjective measures.
In our evaluation, as detailed in Tab.~\ref{tab:evalres}, the proposed method demonstrates advantages over the baseline. 
Specifically, it achieves a Mean Opinion Score (MOS) of 3.71 for audio quality, notably higher than the baseline's 3.61. Additionally, text-to-speech (TTS) models achieve a higher MOS of 3.86, suggesting an upper limit for our accent conversion model. For accent reduction, our method also excels with a score of 1.90 in the Accentness Test, markedly lower than the baseline's 1.99. Meanwhile, speaker similarity ratings were comparable for both methods, as illustrated in Fig.~\ref{fig:speaker_prefer}.
The preference scores of our method and the baseline are close, indicating no significant difference. This is expected as both methods derive from the same TTS model.
\subsection{Ablation Study}
\label{subsec:ab}

First, we enhance the baseline by adding a new loss term, $ \mathcal{L}_{mel^*}$, referred to as \textit{Ablation-1}. The speech representations is not only expected to close to the linguistic representations, but also can generate the same mel-spectrogram as the text representations after passing through the decoder, which achieved an improvement in MOS, from 3.61 to 3.69.

Next, we replace the mel-spectrogram with pre-trained whisper encoder features, relieving the decoupling pressure on the speech encoder. As shown in \textit{Ablation-2} Table~\ref{tab:evalres}, this yields significant improvements in both \textbf{WER} and \textbf{accentness score}, while maintaining a comparable MOS to \textit{Ablation-1}.

Lastly, we also investigate the influence of the pretrain the speech encoder with native speech data before fine-tuning it on accented speech. Compare our method with Ablation-2, we observe a notable reduction in WER from 22.61 to 17.62, while the MOS and accentness scores remain stable. These findings suggest that aligning the linguistic and speech representations through pretraining is beneficial for enhancing the intelligibility of the converted speech.

\section{Conclusion}
\label{sec:typestyle}

In this study, we introduce an innovative reference-free non-autoregressive framework for accent conversion, designed to learn accent-independent linguistic  features that facilitate accent modification in source speech. Our framework leverages TTS-based alignment features to train on non-parallel datasets.
Additionally, we explore the impact of pretraining with native speech data and the incorporation of various acoustic features. Comprehensive evaluations show that these innovations improve audio quality and intelligibility.
\vfill\pagebreak

\bibliographystyle{IEEEbib}
\bibliography{strings,refs}

\end{document}